\documentclass[11pt,a4paper]{article}
\usepackage[hyperref]{acl2020}
\usepackage{times}
\usepackage{latexsym}
\usepackage{xstring}
\usepackage{graphicx}
\usepackage{amsmath}
\usepackage{amsfonts}
\usepackage[normalem]{ulem}
\useunder{\uline}{\ul}{}

\usepackage{booktabs}
\usepackage{microtype}
\usepackage{amsthm}

\newcommand{\guesswhat}{\textit{GuessWhat?!}}
\newcommand{\compguesswhat}{\textit{CompGuessWhat?!}}
\newcommand{\grolla}{\textsc{GroLLA}}

\aclfinalcopy
\def\aclpaperid{1906}

\title{CompGuessWhat?!: A Multi-task Evaluation Framework for Grounded Language Learning}

\author{Alessandro Suglia$^1$, Ioannis Konstas$^1$, Andrea Vanzo$^1$, Emanuele Bastianelli$^1$,\\\textbf{Desmond Elliott$^2$, Stella Frank$^3$,} and \textbf{Oliver Lemon$^1$}\\
  $^1$Heriot-Watt University, Edinburgh, UK \\
  $^2$University of Copenhagen, Copenhagen, Denmark \\
  $^3$University of Edinburgh, Edinburgh, UK \\
  $^1$\texttt{\{as247,i.konstas,a.vanzo,e.bastianelli,o.lemon\}@hw.ac.uk} \\
  $^2$\texttt{de@di.ku.dk} \\
  $^3$\texttt{stella.frank@ed.ac.uk} \\}

\begin{document}

\maketitle
\begin{abstract}
    Approaches to Grounded Language Learning typically focus on a single task-based final performance measure that may not depend on desirable properties of the learned hidden representations, such as their ability to predict salient attributes or to generalise to unseen situations. 
    To remedy this, we present \grolla{}, an evaluation framework for Grounded Language Learning with Attributes with three sub-tasks: 1) Goal-oriented evaluation; 2) Object attribute prediction evaluation; and 3) Zero-shot evaluation.
    We also propose a new dataset \compguesswhat{} as an instance of this framework for evaluating the quality of learned neural representations, in particular concerning attribute grounding. 
    To this end, we extend the original \guesswhat{} dataset by including a semantic layer on top of the perceptual one.
    Specifically, we enrich the VisualGenome scene graphs associated with the \guesswhat{} images with abstract and situated attributes.
    By using diagnostic classifiers, we show that current models learn representations that are not expressive enough to encode object attributes (average F1 of $44.27$). In addition, they do not learn strategies nor representations that are robust enough to perform well when novel scenes or objects are involved in gameplay (zero-shot best accuracy $50.06\%$).
\end{abstract}

\section{Introduction}\label{sec:intro}

\begin{figure}[ht!]
    \centering
    \includegraphics[scale=.3,keepaspectratio]{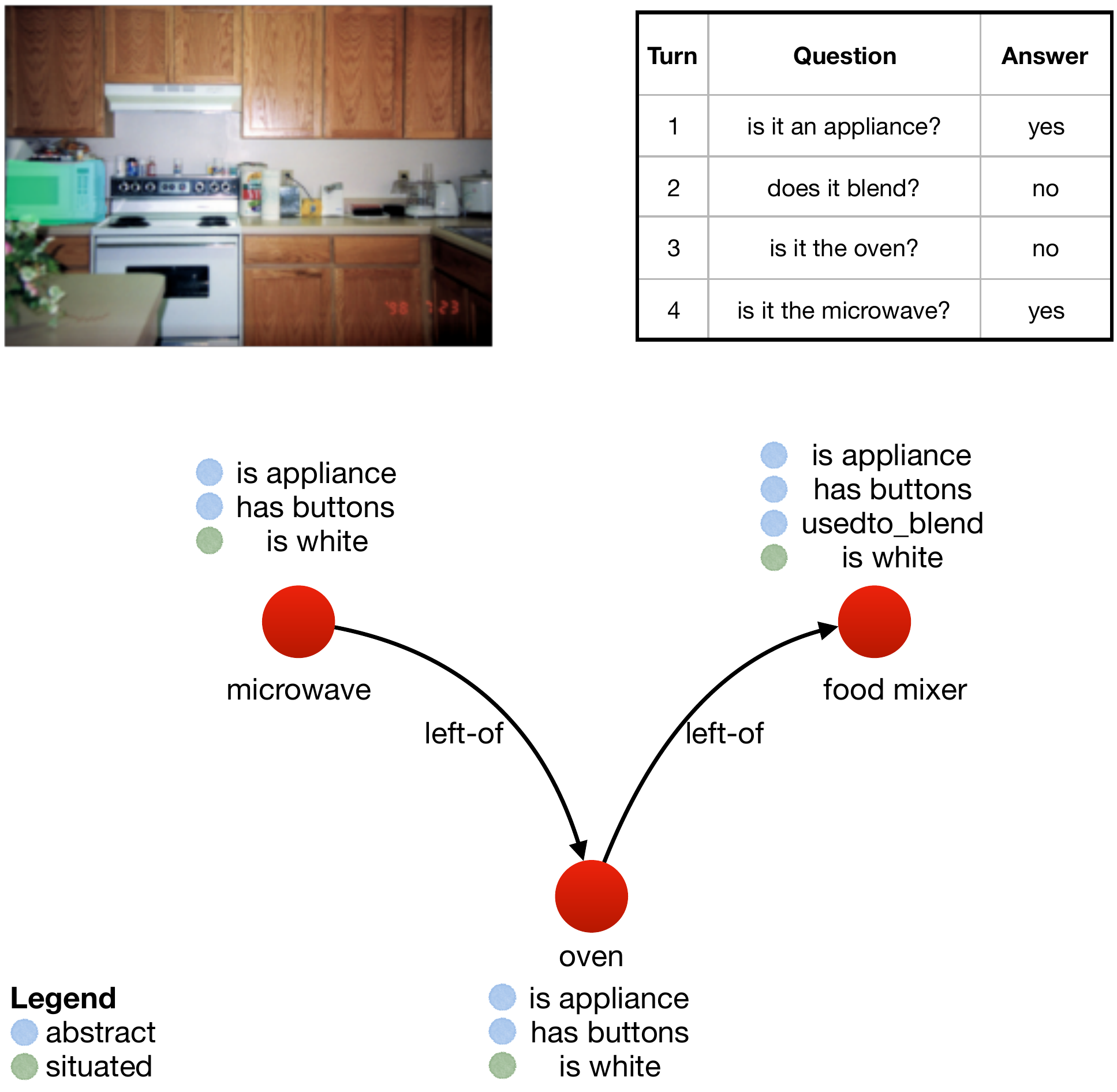}
    \caption{Every gameplay in the \compguesswhat~benchmark has a reference scene that is mapped to a scene graph composed of objects represented in terms of \textit{abstract} and \textit{situated} attributes.}
    \label{fig:compguesswhat}
\end{figure}

Several grounded language learning tasks have been proposed to capture perceptual aspects of language~\cite{shekhar2017foil,hudson2019gqa,Suhr2019:nlvr2,agrawal2018nocaps}. However, the advances in this field have been primarily driven by the final performance measures and less on the grounding capability of the models. In fact, in some cases, high-performance models exploit dataset biases to achieve high scores on the final task~\cite{zhang2016:vqa2,agrawal2016analyzing}. In the literature, several methods have been proposed to analyse what kind of information is captured by neural network representations~\cite{kadar2017representation,belinkov2019analysis}. Most of these works examine the hidden state representations learned by models trained on only textual data. However, many aspects of human semantic representations are grounded in perceptual experience~ \cite{andrews2009integrating,riordan2011redundancy}. 
This paper explores the idea that visually grounded representations ought to be a result of \textit{systematic composition} of grounded representations~\cite{harnad1990symbol}. For instance, the understanding of the word ``microwave" is grounded in perception of objects with specific attributes such as shape, colour, and size -- see Figure \ref{fig:compguesswhat} for an example.
Therefore, investigating whether the representations learned by a model exhibit forms of attribute composition is beneficial for assessing model interpretability and generalisation. 

\begin{figure}[t]
    \centering
    \includegraphics[scale=.3,keepaspectratio]{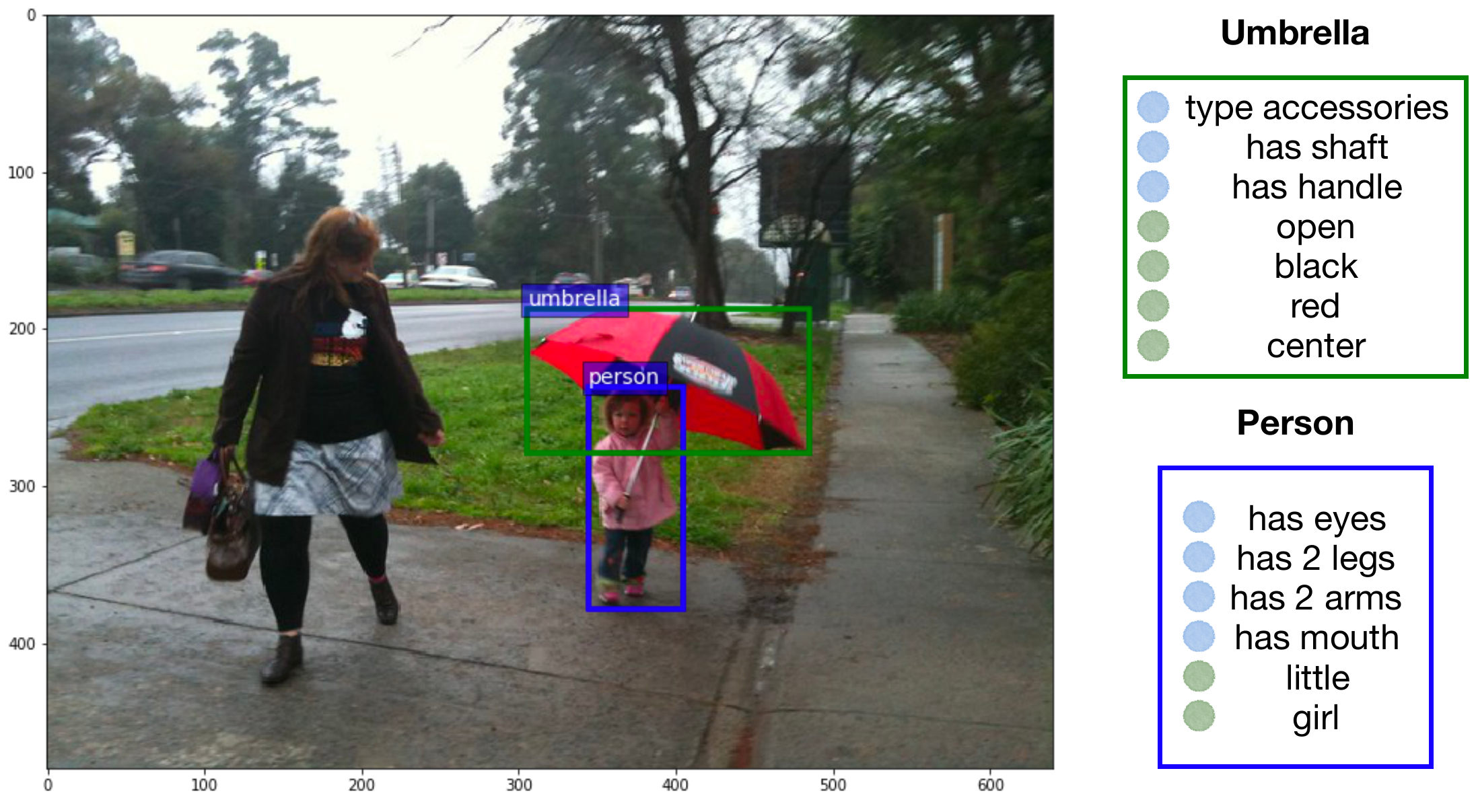}
    \caption{\compguesswhat : Detailed description of the attributes of two different objects in the reference scene. Both the objects have a set of \emph{abstract} attributes (indicated in blue) and a set of \emph{situated} attributes (indicated in green).}
    \label{fig:compguesswhat_attributes}
\end{figure}

In this work, we propose \grolla{} --   a multi-task evaluation framework for Grounded Language Learning with Attributes that expands a goal-oriented evaluation -- based on the standard final task measure, with two auxiliary tasks: 1) Object attribute prediction (AP), and 2) Zero-shot evaluation (ZS). The attribute prediction task is designed to evaluate the extent to which the model's latent representations associated with objects are useful for predicting their attributes. The prediction performance on this task can be related to a \textit{degree of compositionality} of the learned representations. 
We adopt a behavioural, i.e., task-driven, approach to assessing aspects of compositionality for visually grounded representations, whereby the extent to which a representation is compositional depends on: (a) its ability to predict object attributes, and (b) its ability to generalise to novel contributions of object attributes. To support (b), we design a zero-shot evaluation that measures the extent to which the learned representations can be reused in a task involving objects unseen during training.
By optimising for both the final end-goal measure as well as the auxiliary tasks, we aim to drive the design of models that can solve the task more reliably and whose representations are easier to interpret as a result of being a composition of visual attributes.  

This paper presents three main contributions: (1) We define \grolla{} --  a multi-task evaluation framework for grounded language learning that augments the final end-goal measure(s) with auxiliary tasks aimed at assessing the degree of attribute grounding of the model's representations; (2) We propose an instance of this multi-task evaluation framework,  namely \compguesswhat; and (3) We evaluate state-of-the-art models using the \compguesswhat{} dataset. The evaluation shows that models with high performance in the end-goal task are not able to reliably predict the attributes of given objects and do not generalise to examples with unseen object categories.

\compguesswhat~is a benchmark of $65,700$ dialogues (see Section \ref{sec:comp_guesswhat}).
It is based on \guesswhat~\cite{de2017guesswhat} dialogues and enhanced by including object attributes coming from resources such as VISA attributes~\cite{silberer2012grounded}, VisualGenome~\cite{krishna2017visual} and ImSitu~\cite{yatskar2016}.

\section{Evaluation Framework} \label{sec:evaluation_framework}

Our evaluation framework for Grounded Language Learning tasks is based on three different sub-tasks: 1) Goal-oriented evaluation; 2) Object attribute prediction evaluation; 3) Zero-shot evaluation. 

\paragraph{Goal-oriented evaluation}

We evaluate the models according to the multi-modal task that they have to solve, which can generally be categorised as classification or generation. Classification tasks such as Visual Question Answering~\cite{antol2015vqa} or Visual Natural Language Inference~\cite{Suhr2019:nlvr2} involve predicting the correct label for a given example whose performance is measured in terms of predictive accuracy. In generative tasks, such as Image Captioning~\cite{bernardi2016automatic}, the model has to learn to generate a sequence of labels for a given input data whose performance measure is BLEU~\cite{papineni2002bleu}. 

\paragraph{Object attribute prediction evaluation}

We support the goal-oriented evaluation with the attribute prediction auxiliary task related to assessing the degree of compositionality of the representations learned for a specific task.

With an attribute prediction task, we can assess whether the learned representations capture what we think they should, in terms of object attributes, rather than spurious correlations. The idea of using object attributes as an auxiliary task follows from the Characteristic Feature Hypothesis \cite{hampton1979polymorphous} according to which every concept category has a set of \textit{defining features}, which provide a criterion for judging which objects are category members, and which are not. Therefore, the higher the accuracy in the attribute prediction task, the more the representations learned by the model are composed of the set of attributes of the objects.

\paragraph{Zero-shot Evaluation}

Via the attribute prediction task, we can assess the ability of latent representations to recover some of the attributes associated with their object category. Assuming that the model has learned to represent these attributes, we hypothesise that it should solve the original task even when objects that have never been seen during training are involved. 

In our evaluation framework, inspired by other multi-task evaluation frameworks~\cite{wang-etal-2018-glue, mccann2018natural, wang2019superglue, shuster2019dialogue}, we define \textit{Grounded Language Learning with Attributes (\grolla{})} as the final score assigned to the model. It is computed as macro-average of the metrics over all tasks. We define the \grolla{} score for convenience only and we underline the importance of having multiple scores for assessing different model abilities. 
In this work, we present \compguesswhat{} as a dataset implementing this evaluation framework. Thanks to the high overlap between the image set of several datasets~\cite{lu201912}, future work will extend it to other grounded language learning tasks such as image captioning and visual navigation. 

\section{\compguesswhat{} Benchmark}\label{sec:comp_guesswhat}
\subsection{Task Definition}
\compguesswhat{} is an instance of our evaluation framework that is based on a guessing game \cite{steels2015talking}, which can be viewed as a first step in a curriculum of language games for artificial agents. It involves two agents, a scene, and a target object: the Questioner asks questions in order to identify the target object in a scene, while the Oracle knows the target object and has to answer the questions. A multi-word guessing game requires two essential properties for grounded language learning: 1) the ability to \textit{generate} discriminative questions aimed at narrowing down the search space (Natural Language Generation), and 2) the ability to \emph{understand} the information provided so far during the game and exploit it to guess the target object (Natural Language Understanding). 

\subsection{Image Annotations Design}
\compguesswhat{} extends the \guesswhat{} dataset \cite{de2017guesswhat} to promote the study of {\it attribute}-grounded language representations. The original \guesswhat{} dataset is extended with a semantic layer on top of the perceptual layer (i.e., images). This layer consists of a collection of intentional and extensional attributes of the objects in the reference image (Figure \ref{fig:compguesswhat_attributes}). 
We enrich the VisualGenome~\cite{krishna2017visual} scene graphs associated with the \guesswhat~images with several attributes coming from resources such as VISA~\cite{silberer2012grounded} and ImSitu~\cite{yatskar2016}. 
Unfortunately, not all the \guesswhat{} images are included in VisualGenome. We were able to reuse $40.79\%$ of the original \guesswhat{} dialogues for a total of $65,700$ dialogues (additional information can be found in the related Appendix \ref{appendix:comp_guesswhat_dataset}). By relying on this set of attributes, we define an \textit{attribute prediction evaluation} to assess the extent to which the learned neural representations can encode the attributes specified during the dialogue. In order to determine the generalisation power of the learned representations and their ability to be transferred, we propose a novel \textit{zero-shot learning} set of reference games involving target object belonging to an unseen object category. The dataset and the code associated with this paper can be found online\footnote{\url{https://compguesswhat.github.io}}.

\paragraph{Psycholinguistically-motivated attributes} We extend the set of attributes for every object category in MSCOCO with psycholinguistically-motivated semantic representations based on the McRae Norms~\cite{mcrae2005semantic} developed by \citet{silberer2012grounded}. 
We use only the subset of so-called \emph{abstract} attributes, and ignore attributes from the original set
that can change depending on the reference image (e.g., ``shape", ``texture", etc.). We use the WordNet synset identifier (e.g., ``person.n.01") associated with a given MSCOCO category (e.g., ``person'') to automatically associate its corresponding abstract attributes with a specific object instance. However, very often several VisualGenome objects have a synset associated with a class that is a hyponym of the MSCOCO category synset. Therefore, we rely on the Wu-Palmer similarity~\cite{wu1994verbs} to find the best match between the VisualGenome synset and the MSCOCO category synset (with a similarity threshold of $0.75$ chosen by using as reference the distance between the synset of \textit{person} and \textit{woman}). The intuition behind this heuristic is that we assume that a hyponym will inherit the abstract attributes of its hypernym.

\paragraph{Affordances \& Behaviours} We extract the semantic roles associated to specific object categories using the \textit{ImSitu}~dataset \cite{yatskar2016}, in order to include affordances and behaviours associated with every object category. 
An object category is associated with a \textit{behaviour} every time it appears as the \emph{agent} of a given predicate. For instance, ``the food mixer [agent] blends fruit", where the \textit{behaviour} is the food mixer's ability to blend something. We also consider \textit{affordances} associated with a given category and divide them into two categories: 1) \emph{can\_be}, every predicate having the object category as \emph{item}, \emph{coagent}, \emph{vehicle} semantic role;  2) \emph{used\_to}, every predicate having the object category as \emph{tool}, \emph{heatsource}, \emph{object}. For example, in the statement ``the person opens the oven [item]" an \textit{affordance} can be intended as the fact that an oven \emph{can be} opened. These attributes extend the set of \emph{abstract} attributes. The abstract attributes do not depend on the reference image so they can be reused in other contexts as well.

\paragraph{Situated attributes} Since the images contained in \guesswhat{} come from the MSCOCO dataset (see Figure \ref{fig:compguesswhat} for an example), some of them are included in the VisualGenome~\cite{krishna2017visual} dataset, which is composed of rich scene graphs for every image. In particular, we verified that $27,155$ images from the \guesswhat{} dataset are also contained in VisualGenome. However, due to the presence of possible visual elements, the VisualGenome images are not the same as the MSCOCO ones. We use a heuristic approach based on both Intersection over Union (IoU) and language-only features to match the object bounding boxes between the two images. We report more details about the algorithm in Appendix \ref{appendix:vg_object_mapping}. The set of object attributes from VisualGenome (attribute types, colour, size, etc.) and location/positional attributes (one of top/bottom/left/right/centre, based on bounding box location) make up the \textit{situated attributes}, which are specific to the reference image.

As a final step, due to the image mismatch, we decided to include the original \guesswhat{} object annotations in the VisualGenome graph in case a \guesswhat{} object cannot be mapped to a VisualGenome one. By doing this, we have access to the MSCOCO category of the object from which we can recover all its abstract attributes.

\section{\compguesswhat{} Evaluation} \label{sec:evaluation}

\subsection{Guesser accuracy evaluation}

We consider the guesser accuracy metric (in gameplay mode\footnote{A gameplay involves three trained models that generate dialogues given a pair of (image, target object).}) from the \guesswhat{} dataset for our goal-oriented evaluation. It measures how many times the guesser model can select the correct target object among the candidate objects, given the dialogue generated so far. Due to the importance of this language game for NLU and NLG model skills, we decide to keep the guesser accuracy as a reference metric to assess the ability of the questioner to play the game. However, unlike the original dataset evaluation, we make sure that the score is evaluated ignoring duplicated dialogues.\footnote{In the test dataset multiple conversations are associated with the same (image, target object) pair. Therefore, we want the pair (image, target object) to be considered only once in the accuracy evaluation.} 

\subsection{Attribute Prediction Evaluation}

In a sequential guessing game like the one in Figure~\ref{fig:compguesswhat}, 
we regard the representation for the last turn of the dialogue as a composition or aggregation of all the attributes specified so far. Therefore, we can use it to predict with high accuracy the attributes associated with a specific target object because it should encode the information needed to correctly discriminate the target from all the other objects in the scene. In the dialogue of Figure~\ref{fig:compguesswhat}, when the model generates a representation for the last turn of the conversation (i.e., ``Q: Is it the microwave? A: Yes''), it should encode the fact that ``it is an appliance'', ``it is not the oven'' and ``it is the microwave'', allowing the agent to guess the target object correctly.

By playing several guessing games that have a microwave as the target object, the agent should learn a representation of microwave that is expressive enough to correctly discriminate a microwave from all the other objects in a scene. In this setup we are not assuming that the model has a \textit{single} representation for the concept of microwave; rather the concept of microwave develops from aggregating multimodal information related to microwaves across the situations in which the object is experienced \cite{barsalou2017cognitively}. In the context of \compguesswhat{}, every successful dialogue involving a microwave as the target object will be considered as an \emph{experience}.

We are interested in understanding whether the dialogue state representation generated by a neural model for the last turn of the dialogue can encode the attributes of the target object specified during the dialogue. To do so, we define four attribute prediction tasks. For every target object we predict the corresponding vector composed of: 1) \textit{abstract} attributes only (\textit{A}); 2) situated attributes only (\textit{S}), 3) the union of abstract and situated attributes (\textit{AS}), and 4) location attributes (\textit{L}) such as \textit{center}, \textit{top}, \textit{bottom}, \textit{right} and \textit{left}. After training the model on the original \guesswhat{} dataset, we can generate dialogue representations corresponding to all the \compguesswhat{} successful games. Then, we can train a \textit{diagnostic classifier} that predicts the attributes associated with a given object category using the dialogue hidden representation generated for a given game as features. We hypothesise that a model that has learned grounded representations that are expressive enough to correctly guess the target object should retain the relevant features to predict its attributes. 

We treat the attribute-prediction problem as a multi-label classification task. We implement our diagnostic classifier $\Phi$ as a linear transformation parameterised by a weight matrix $\mathbb{R}^{d_d \times d_a}$ (where $d_d$ is the dialogue hidden state size and $d_a$ is the number of attributes to be predicted) followed by a sigmoid activation function. We use a sigmoid activation function because it models a Bernoulli distribution. The diagnostic classifier outputs $d_a$ logits where each of them models the probability $P(y_k = 1 | \mathbf{d})$ (where $\mathbf{d}$ is dialogue state representation), one for each attribute $y_k$ to be predicted. To mitigate a possible class-imbalance problem, we apply a filtering strategy to remove underrepresented attributes from our attribute set, which is a similar technique used to deal with out-of-vocabulary words. We also decided to avoid using class-weighting so that we could evaluate the power of the learned representations with simple linear classifiers as done in previous work using probing classifiers~\cite{belinkov2019analysis}. Please refer to Appendix \ref{appendix:attribute_prediction} for details about the procedure to derive the reference set of attributes.

We use the \compguesswhat{} dataset split as the reference for our training and evaluation setup: we train the diagnostic classifiers on \compguesswhat{} gold training dialogues and evaluate their performance on the test dialogues using the validation set dialogues for early stopping. We consider Precision, Recall, and F1-measure for multi-label classification~\cite{sorower2010literature} (computed as \textit{macro-average}) and evaluate them with $0.5$ as the threshold value for the sigmoid activation function (selected after considering the models performance using threshold values of $0.75$ and~$0.9$). We report additional details in Appendix~\ref{appendix:attribute_prediction}. 

\subsection{Zero-shot Evaluation}
 
Assuming that the model has learned to compose concepts during the turns of the dialogue, we hypothesise that it should also be able to \emph{use} these representations to play games involving target objects that belong to categories that have never been seen before. For example, humans can discriminate between a \emph{dolphin} and a \emph{dog} even though they might not know what it is called. The measure presented in this section has the potential to demonstrate whether current models lack the ability to systematically generalise to new instances that are composed of attributes learned during training. 

In order to assess the true generalisation power of the trained agents, we define a zero-shot learning scenario based on the \textit{nocaps} dataset images \cite{agrawal2018nocaps}. The \textit{nocaps} dataset is composed of 3 evaluation splits: 1) \textit{in-domain}: annotated objects belong to MSCOCO categories only; 2) \textit{near-domain}: contains a mixture of MSCOCO and OpenImages objects; 3) \textit{out-of-domain}: contains only OpenImages object categories. Since the number of categories in the original \guesswhat{} dataset (80) is lower than the number of categories in the Open Images dataset (660) -- contained in \textit{nocaps} -- there are many categories that are never seen during training. Therefore, we can create zero-shot learning games by considering a target object for the game whose category has never been seen during training. We define an automatic procedure to generate the set of reference games for the zero-shot learning setup using the nocaps images. We split the nocaps images into near-domain or out-of-domain. An image is considered near-domain if it contains at least one object whose category belongs to MSCOCO. In contrast, we consider the image out-of-domain if it does not contain any MSCOCO category objects. This procedure generates a dataset of $19,179$ near-domain reference games and $18,672$ out-of-domain reference games.
More details about the automatic procedure as well as the resulting reference set of games can be found in Appendix \ref{appendix:zero_shot_eval}. 
As a last step of our evaluation framework, we evaluate the performance of the state-of-the-art models in the zero-shot gameplay setup. For this task, the trained models need to interact with each other and generate dialogues given the pair (image, target object). As an evaluation metric for this task, we consider gameplay guesser accuracy for the \textit{near-domain}~(ND-Acc) and \textit{out-of-domain}~(OD-Acc) reference games. 

\begin{table*}[t]
\centering
\begin{tabular}{lcccccccc}
\toprule
                                      & \textit{\textbf{Gameplay}} & \multicolumn{4}{c}{\textit{\textbf{Attribute Prediction}}}           & \multicolumn{2}{c}{\textit{\textbf{Zero-shot Gameplay}}} & \multicolumn{1}{l}{\textit{\textbf{GroLLA}}} \\
 & \textbf{Accuracy}          & \textbf{A-F1}   & \textbf{S-F1}  & \textbf{AS-F1}  & \textbf{L-F1}   & \textbf{ND-Acc}             & \textbf{OD-Acc}             & \multicolumn{1}{l}{} \\   
\cmidrule(lr){2-2} \cmidrule(lr){3-6} \cmidrule(lr){7-8} \cmidrule(lr){9-9}
Random                                & 15.81\%                    & 15.1          & 0.1           & 7.8           & 2.8          & 16.9\%                     & 18.6\%                     & 13.3                                       \\
\midrule
GloVe                                 & -                          & 34.6          & 29.7          & 36.4          & 33.6          & -                           & -                           & -                                             \\
ResNet                                & -                          & 24.5          & 31.7         & 27.9          & 43.4          & -                           & -                           & -                                             \\
\midrule
GDSE-SL-text                          & -                          & 57.0            & 45.3          & 57.5          & 46            & -                           & -                           & -                                             \\
GDSE-CL-text                          & -                          & 56.9         & 45.0            & 57.3          & 45            & -                           & -                           & -                                             \\
\midrule
DeVries-SL                            & 41.5\%                     & 46.8          & 39.1          & 48.5          & 42.7          & 31.3\%                     & 28.4\%                     & 38.5                                       \\
DeVries-RL                            & 53.5\%                     & 45.2          & 38.9          & 47.2          & 42.5          & \textbf{43.9}\%                     & \textbf{38.7\%}            & 46.2                                       \\
\midrule
GDSE-SL                               & 49.1\%                     & \textbf{59.9} & 47.6 & \textbf{60.1} & \textbf{48.3} & 29.8\%                     & 22.3\%                     & 43.0                                       \\
GDSE-CL                               & \textbf{59.8\%}            & 59.5          & 47.6          & 59.8            & 48.1          & 43.4\%            & 29.8\%                     & \textbf{50.1}                              \\
\bottomrule
\end{tabular}
\caption{Results for state-of-the-art models on the \compguesswhat{} suite of evaluation tasks. We assess model quality in terms of \textit{gameplay} accuracy, the \textit{attribute prediction} quality, measured in terms of F1 for the \textit{abstract}~(A-F1), \textit{situated}~(S-F1), \textit{abstract+situated}~(AS-F1) 
and \textit{location}~(L-F1) prediction scenario, as well as \textit{zero-shot learning gameplay}. The final score \grolla{} is a macro-average of the individual scores. We use the models \texttt{GloVe}, \texttt{ResNet} and \texttt{GDSE-*-text} only as a baseline for the attribute prediction tasks.}
\label{tab:summary_results}
\end{table*}

\section{Results: Model Evaluation using \compguesswhat}
\paragraph{Guesser accuracy}
We evaluate the \texttt{GDSE} and \texttt{DeVries} models in gameplay mode using the set of reference games provided in \compguesswhat{}. As shown in Table \ref{tab:summary_results}, the results are in line with the performance of the models on the original \guesswhat{} dataset~\cite{de2017guesswhat,shekhar2019beyond} confirming that our filtering strategy did not affect the complexity of the task.

\paragraph{Attribute Prediction}

We use the \compguesswhat{} benchmark to compare several dialogue state representations:

\noindent
\texttt{DeVries-SL}: the representation learned by the Questioner model presented in \cite{de2017guesswhat} that generates the question tokens conditioned on the image features and is trained using Supervised Learning (SL).

\noindent
\texttt{DeVries-RL}: the representations learned by the Questioner model presented in \cite{de2017guesswhat}, fine-tuned using the Reinforcement Learning procedure proposed in \cite{strub2017end}.

\noindent
\texttt{GDSE-SL}: the \textit{grounded dialogue state} learned by a seq2seq model trained using the multi-task Learning procedure in \cite{shekhar2019beyond}.

\noindent
\texttt{GDSE-CL}: the \textit{grounded dialogue state} learned by the Questioner model used in \texttt{GDSE-SL}, fine-tuned with the Collaborative Learning procedure presented in \cite{shekhar2019beyond}. 

\noindent
\texttt{GDSE-SL-text}: the learned LSTM~\cite{hochreiter1997lstm} dialogue encoder of the \texttt{GDSE-SL} model.

\noindent
\texttt{GDSE-CL-text}:\footnote{We could use the dialogue encoder of the \texttt{GDSE} models only due to their modular architecture. It was not possible to properly separate the dialogue encoder from the visual representation in the \texttt{DeVries} models.} the learned dialogue encoder of the \texttt{GDSE-CL} model.

In order to control for possible bias in our task, we consider unimodal~\cite{thomason2019shifting} as well as random attribute predictors:

\noindent
\texttt{GloVe}: a dialogue is represented as the average of the GloVe embeddings associated with each word~\cite{pennington2014glove}.

\noindent
\texttt{ResNet}: uses the latent representation of the reference scene generated by a ResNet152 as proposed in \citet{shekhar2019beyond}.

\noindent
\texttt{Random}: samples $d_a$ scores from $U(0,1)$ where samples are independent from each other. We incorporate this baseline as a lower bound performance on the attribute prediction task.

With the AP task, we try to answer the following question: ``Do the representations associated with the target object encoding provide useful information that can be exploited to predict the object attributes correctly?" We assume that, due to the nature of the \compguesswhat{} games, the final dialogue state representation should encode relevant features of the target object. So, a high gameplay accuracy should correlate with a high AP score. Table~\ref{tab:summary_results} summarises the results of the attribute prediction task evaluated on the \compguesswhat{} test games. As the average best model performance was only $44.27$, far from ceiling, our hypothesis is only partially supported. In particular, the models having the highest guesser accuracy, \texttt{GDSE-CL} and \texttt{GDSE-SL}, seem to learn better representations than unimodal baselines \texttt{GloVe} and \texttt{ResNet}, confirming the importance of multi-modal training for this task. There is also a gap in performance  between the \texttt{GDSE} and \texttt{DeVries} models. This might be related to the multi-task learning strategy used by \texttt{GDSE} models that favours the emergence of more expressive representations than the ones learned by \texttt{DeVries} models which are trained in isolation. By comparing the enhanced versions \texttt{GDSE-CL} and \texttt{DeVries-RL} with the less sophisticated ones, \texttt{GDSE-SL} and \texttt{DeVries-SL}, respectively, we observe that, despite their higher guesser accuracy, these models do not have any advantage in terms of the AP task. We believe that this is because the Collaborative training strategy (for \texttt{GDSE-CL}) and Reinforcement Learning (for \texttt{DeVries-RL}) are optimising end-goal performance while sacrificing the expressiveness of the representations. Finding a way to encode \textit{task-specific} representations and generalise them to learn \textit{abstract representations} becomes an important research direction to improve on this task. 

As an additional ablation, we compared the representations learned by the LSTM module used by \texttt{GDSE} to encode the dialogue (\texttt{GDSE-*-text}) with their \textit{grounded dialogue state} counterpart. Differences in terms of F1 are minimal, confirming that the heavy lifting is done by the textual representations and it is not clear how well the \textit{grounded dialogue state} retains the visual information. Another confirmation of this issue is provided by the results in terms of location attributes prediction. Performance in this task for all the models is around~$40$ meaning that both VGGNet and ResNet features (used for \texttt{DeVries} and \texttt{GDSE}, respectively) are not able to recover fine-grained object information. This result sheds light on the ability of these models to ground the textual data in perceptual information of the reference scene. We believe that models should be able to \textit{co-ground} one modality with the other and, as a result, learn more expressive grounded representations. 

\paragraph{Zero-shot Evaluation}

\begin{figure*}[t]
    \centering
    \includegraphics[scale=.19,keepaspectratio]{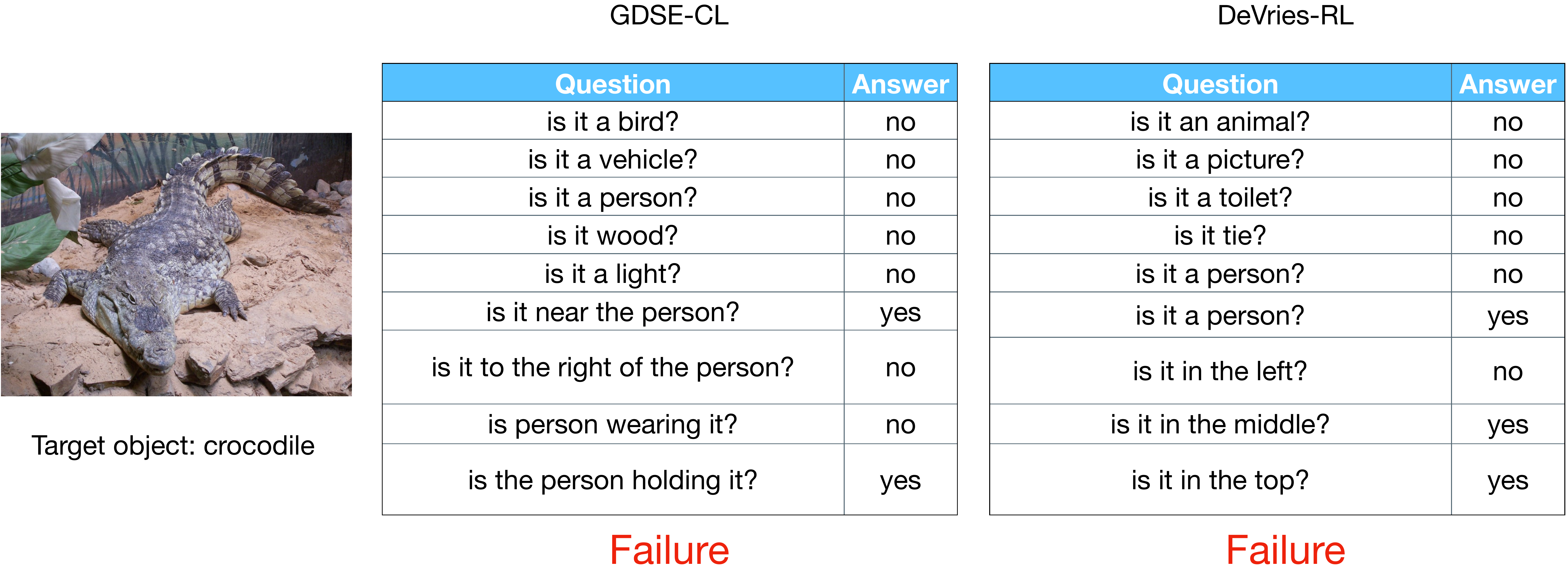}
    \caption{Example gameplay in the out-of-domain scenario of the two best performing systems GDSE-RL and DeVries-RL. The models have to play the game considering the \textit{crocodile} as target object. This is a zero-shot scenario because the object category \textit{crocodile} is not among the MSCOCO categories.}
    \label{fig:gameplay_zs}
\end{figure*}

Results are summarised in Table~\ref{tab:summary_results}; the most striking observation is that all models struggle with this dataset (guesser accuracy is barely above $40$),
although arguably humans would be able to solve the task despite their unfamiliarity with a specific object. 
Indeed, in this zero-shot scenario, reusing previously learned attributes that are shared among the objects or leveraging \textit{mutual exclusivity}~\cite{markman1988children} would result in a successful gameplay. 

Even the most accurate model in the \compguesswhat{} guesser evaluation performs poorly in this zero-shot setup (see Figure \ref{fig:gameplay_zs} for an example). We attribute this drop in performance to the way that these models represent objects. 
In particular, they all rely on \textit{category embeddings}, i.e., latent representations associated to specific object categories (refer to \cite{shekhar2019beyond,de2017guesswhat} for more details). In the case of ZS evaluation, when an object is unknown, its category embedding is also not available. This is true for both \texttt{DeVries} and \texttt{GDSE} models; it seems that \texttt{GDSE} models suffer more than \texttt{DeVries} models possibly due to overfitting. On the other hand, we aim to learn object representations which are not associated with manually-provided categories but are obtained by playing the game and that encode both abstract and situated attributes. 

Once again, we find that models optimised using Reinforcement Learning seem to learn a better game strategy that results in higher performance on both near-domain and out-of-domain games. To better understand the quality of the generated dialogues, we classify each type of question according to a pre-defined set of types based on 
\cite{shekhar2019beyond} (please refer to Appendix \ref{appendix:dialogue_eval} for a detailed description and a detailed summary of the evaluation results). We noticed that the \texttt{DeVries} models generate dialogues with $70\%$ of their turns comprising ``location'' questions (e.g., ``is it the person on the right?") compared to $20\%$ for \texttt{GDSE} models. 
We argue that to tackle zero-shot scenes, a model should instead learn features useful to discriminate the target object without relying on locations. 
Of course, in some reference scenes, location questions are still useful attributes used by humans when playing the game. In addition, asking location questions is an effective strategy because the Oracle has access to positional information that can be used to provide reliable answers but does not have any category embeddings for the target object.

\section{Related Work} \label{sec:related_work}

\paragraph{Text-only Multi-task Evaluation}
In recent years progress in NLP has been driven by multi-task evaluation datasets proposed to mitigate the biases of task-specific datasets~\cite{wang-etal-2018-glue, mccann2018natural, wang2019superglue}.
Despite their multi-task nature, these datasets focus on text-only data making the resulting models unable to learn meaning representations which are grounded in perceptual experience~\cite{andrews2009integrating,riordan2011redundancy}. Another downside is that these benchmarks focus only on end-goal metrics, i.e., are not informative on what the model has learned. Going beyond the end-goal metric is fundamental for designing models that are more generalisable and interpretable. By introducing the attribute prediction task in our framework, we assess whether the learned representations are expressive enough to predict the attributes of relevant objects in the scene. Also, we propose a \textit{zero-shot evaluation} where the model has to generate predictions for examples that have never been seen during training, thus providing a way to understand the generalisation power of the learned representations.

\paragraph{Grounded Language Learning Evaluation}

Several grounded language learning tasks have been proposed in the literature that can be divided into discriminative~\cite{shekhar2017foil,hudson2019gqa,Suhr2019:nlvr2} and generative grounded language learning tasks~\cite{xu2015show,agrawal2018nocaps}. Recent works proposed models trained in a multi-task fashion by exploiting several language/vision tasks. The \textit{dodecaDialogue} task~\cite{shuster2019dialogue} proposes twelve dialogue tasks, among which there are two language/vision tasks in which the agent has to generate a response for a given context. Other works try to exploit multi-task learning to improve on single-task model performance in discriminative tasks~\cite{pramanik2019omninet, lu201912}. Unfortunately, implementing multi-task learning using different datasets results is cumbersome~\cite{subramanian2018learning}. We propose an evaluation framework that can be applied in the context of a single task and dataset (e.g.~\guesswhat{}) that allows to understand the extent to which the model can learn useful representations for the task at hand.

Inspecting the learned representations is important because, due to biases in the datasets, models might learn spurious correlations between input and output rather than actual grounding capabilities. For instance, in Visual Question Answering, questions starting with ``What colour are" have  ``white" as a correct answer 23\% of the time; models learn to memorise this sort of association rather than using the visual information~\cite{zhang2016:vqa2,agrawal2016analyzing}. This issue calls for a model evaluation aimed at inspecting the model representations as well as how these representations are used. The GQA~\cite{hudson2019gqa} dataset goes in this direction. 
It presents a Visual Question Answering dataset where images are supported by rich semantic annotations in the form of scene graphs. The GQA task requires the model to select an answer among a set of candidates. 

However, we advocate the importance of tasks that involve both Natural Language Understanding (NLU) and Natural Language Generation (NLG) skills in a curriculum for grounded language learning. There are significant differences concerning the proposed auxiliary tasks as well. First of all, GQA's tasks are specifically designed around the VQA tasks to make sure that the model is \emph{consistent} and \emph{plausible}. It does not however tell us what the model's learned representations are encoding. 

We propose the AP task as a diagnostic task aimed at better understanding the learned neural representations~\cite{belinkov2017analyzing, conneau2018probing, peters2018dissecting, tenney2019you}. In addition, going beyond simple object classification is considered beneficial for vision systems~\cite{farhadi2009describing} because it allows generalisation across object categories, not just across instances within a category. However, we believe that to truly assess the generalisation ability of a model, object attributes have to be used for the downstream task, which is not necessarily needed in object classification tasks. With the ZS evaluation, we investigate the ability of the models to exploit more fine-grained visual attributes which is important for models able to learn from few examples and easily transfer to new domains.

\paragraph{Compositionality Evaluation} 

\citet{andreas2019measuring} presents a method to estimate the degree of compositionality of neural representations by using an \textit{oracle} compositional model aware of the compositional structure (i.e., a derivation) of the input data. Building a reference oracle is easy for synthetic scenes (as in \citet{andreas2019measuring}) but is a significant challenge for real-world scenes. Previous work has studied compositionality in real-world scenes for visual concept composition ~\cite{misra2017red} and image captioning ~\cite{nikolaus-etal-2019-compositional}. In our benchmark \compguesswhat{}, we use real-world scenes from the MSCOCO~\cite{lin2014microsoft} and OpenImages~\cite{juznetsova2018:openimages} datasets. Our AP task is related to measuring compositionality. It relies on image annotations in the form of intensional and extensional attributes as a reference structure for the objects in the scene. 

\section{Conclusions \& Future Work}

We proposed \compguesswhat{} as an implementation of \grolla{}, a multi-task evaluation framework for Grounded Language Learning with Attributes.
We found that the best performing model achieves a \grolla{} score of $50.06\%$; notably this model's out-of-domain accuracy is under $30\%$, as compared to the human performance on the original \guesswhat{} dataset of $90.2\%$~\cite{de2017guesswhat}. Clearly, even models with high in-domain gameplay success rates still have difficulty generalising to new scenarios. In the following, we discuss insights gained from the evaluation and new research directions for this task.

The attribute prediction task shows that model representations are not able to accurately recover attribute representations. We argue that this result calls for new approaches to exploiting and representing textual and visual data. We believe that models should be equipped with a \emph{co-grounding} operator that fuses the textual and visual modalities. For instance, in the context of \compguesswhat, it would be used to learn a representation for the current turn that is influenced by \emph{both} the language and visual modality. 
\compguesswhat{} requires models to learn to combine the co-grounded information provided for every turn. Therefore, we propose that \compguesswhat{} represents a benchmark dataset for evaluating the design of such an \textit{attribute compositionality operator} that would be a possible implementation of compositionality à la~\citet{barsalou2017cognitively}. 

In this work, we have shown how our multi-task evaluation framework can be be applied to \guesswhat{}. However, the same framework could be applied to other multi-modal tasks. For example, in image captioning, the goal-oriented evaluation would be the textual similarity metrics (e.g.~BLEU); the attribute-prediction task would use the decoder representation to predict the attributes of the objects in the image~\cite[e.g.]{elliott2017imagination}; and the zero-shot setting could leverage the nocaps dataset \cite{agrawal2018nocaps}. Likewise, in the Vision-and-Dialog navigation task~\cite{thomason2019vision}, the goal-oriented evaluation is the navigation task; attribute prediction is based on predicting the attributes of the hidden object when the agent decides it is in the correct room, and the zero-shot setting could evaluate model performance on novel combinations of rooms and object types.

Finally, from the evaluation presented here, it emerges that these models learn task-specific representations that do not generalise to unseen object categories. We hope that \grolla{} and the \compguesswhat{} data will encourage the implementation of learning mechanisms that fuse task-specific representations with more abstract representations to encode attributes in a more compositional manner. In addition, we will use the \compguesswhat{} image annotations to design a visual grounding evaluation to assess the ability of the model to attend to the correct objects during the turns of the dialogue.

\section*{Acknowledgements}

We thank Arash Eshghi and Yonatan Bisk for fruitful discussions in the early stages of the project.

\bibliographystyle{acl_natbib}
\bibliography{acl2020}

\clearpage

\appendix

\section{Appendices}
\label{sec:appendix}

\subsection{\compguesswhat~Dataset}\label{appendix:comp_guesswhat_dataset}

We extend the \guesswhat{} dataset \cite{de2017guesswhat} to promote the study of {\it compositional} grounded language representations. The original \guesswhat{} dataset has been enhanced by including a semantic layer on top of the purely perceptual one (i.e., images). In particular, we enrich the VisualGenome~\cite{krishna2017visual} scene graphs associated with the \guesswhat~images with several attributes coming from resources such as VISA~\cite{silberer2012grounded} and ImSitu~\cite{yatskar2016}. As shown in Table \ref{tab:image_mapping_vg_gw} not all the \guesswhat~images are included in VisualGenome. We were able to reuse $40.79\%$ of the original \guesswhat~dialogues for a total of $65,700$ dialogues as summarised in Table \ref{tab:dataset_statistics}.

\begin{table}[ht!]
\centering
\small
\begin{tabular}{|l|l|l|}
\hline
\textbf{Split} & \textbf{GuessWhat?! images} & \textbf{Mapped images} \\ \hline
Train          & 46794                       & 19117                  \\ \hline
Validation     & 9844                        & 4049                   \\ \hline
Test           & 9899                        & 3989                   \\ \hline
\end{tabular}
\caption{Statistics of the mapping between GuessWhat?! images and VisualGenome images.}
\label{tab:image_mapping_vg_gw}
\end{table}

\begin{table*}[ht!]
\small
\begin{tabular}{|l|l|l|l|l|l|l|l|}
\hline
\textbf{Split} & \textbf{\begin{tabular}[c]{@{}l@{}}\# \guesswhat\\ dialogues\end{tabular}} & \textbf{\begin{tabular}[c]{@{}l@{}}\# \compguesswhat\\ dialogues\end{tabular}} & \textbf{\begin{tabular}[c]{@{}l@{}}Vocab.\\ size\end{tabular}} & \textbf{\begin{tabular}[c]{@{}l@{}}Avg. dialogue\\ length\end{tabular}} & \textbf{\begin{tabular}[c]{@{}l@{}}Successful\\ dialogues\end{tabular}} & \textbf{\begin{tabular}[c]{@{}l@{}}Failed\\ dialogues\end{tabular}} & \textbf{\begin{tabular}[c]{@{}l@{}}Incomplete\\ dialogues\end{tabular}} \\ \hline
Train          & 113221                                                                   & \begin{tabular}[c]{@{}l@{}}46277 \\ (40.92\%)\end{tabular}               & 7090                                                           & 5.128                                                                   & \begin{tabular}[c]{@{}l@{}}84.06\%\\ (38901)\end{tabular}               & \begin{tabular}[c]{@{}l@{}}10.35\%\\ (4790)\end{tabular}            & \begin{tabular}[c]{@{}l@{}}5.59\%\\ (2586)\end{tabular}                 \\ \hline
Valid          & 23739                                                                    & \begin{tabular}[c]{@{}l@{}}9716\\ (41.02\%)\end{tabular}                 & 3605                                                          & 5.106                                                                   & \begin{tabular}[c]{@{}l@{}}83.97\%\\ (8159)\end{tabular}                & \begin{tabular}[c]{@{}l@{}}11.03\%\\ (1069)\end{tabular}            & \begin{tabular}[c]{@{}l@{}}5.03\%\\ (488)\end{tabular}                  \\ \hline
Test           & 23785                                                                    & \begin{tabular}[c]{@{}l@{}}9619\\ (40.44\%)\end{tabular}                 & 3552                                                           & 5.146                                                                   & \begin{tabular}[c]{@{}l@{}}84.10\%\\ (8090)\end{tabular}                & \begin{tabular}[c]{@{}l@{}}10.74\%\\ (1034)\end{tabular}            & \begin{tabular}[c]{@{}l@{}}5.14\%\\ (495)\end{tabular}                  \\ \hline
\end{tabular}
\caption{Comparison between the original \guesswhat~dataset and \compguesswhat~dataset. We report the percentage of dialogues that we retain after the filtering procedure based on the VisualGenome images.}
\label{tab:dataset_statistics}
\end{table*}

\subsection{VisualGenome object mapping} \label{appendix:vg_object_mapping}
VisualGenome images are not exactly the same in terms of shape and content as the ones in MSCOCO. This is due to the presence of possible visual elements (i.e., banners) that are in the VisualGenome version of the image and are not in the MSCOCO one. This complicates the object mapping procedure used to link together abstract attributes and attributes coming from VisualGenome. 
As a first step, the procedure finds the largest VisualGenome bounding box with an IoU greater than $0.5$. If there is not one, it looks for the largest VisualGenome bounding box with an IoU which is not close to $0$ (with a tolerance of $0.05$) and whose category is similar to the one of the reference MSCOCO one (where `similar' is measured according to the Jaccard index between the corresponding category tokens). Whenever the MSCOCO object bounding box cannot be mapped to one of the VisualGenome bounding boxes, we assume that we do not have access to the situated attributes and we use the abstract attributes associated to its MSCOCO category only. 

\subsection{Diagnostic Classifiers for Attribute Prediction}\label{appendix:attribute_prediction}

For the attribute prediction task we apply a filtering procedure on the attribute set that will be used for training. In particular, we ignore all the attributes that belong to the abstract attribute category whose frequency is less than $100$ (resulting in a set of attributes equal to $1997$) and we ignore all the situated attributes whose frequency is less than $2$ (resulting in a set of attributes equal to $4085$).

For the attribute-prediction task we define a probing classifier $\Phi$ as a linear transformation parameterised by a weight matrix $\mathbb{R}^{d_d \times d_a}$ (where $d_d$ is the dialogue hidden state size and $d_a$ is the number of attributes to be predicted) followed by a sigmoid activation function. The number of input dimensions $d_d$ depends on the model hidden state representations. We report in Table \ref{tab:hidden_states} the corresponding hidden state sizes for all the evaluated models. The output size $d_a$ depends on the attribute set that we intend to consider. When situated attributes are considered $d_a = 6082$,  $d_a = 1997$ for abstract attributes, $d_a = 5$ for location attributes and $d_a = 4085$ for situated-only attributes.

\begin{table}[ht!]
\centering
\begin{tabular}{|l|l|}
\hline
\textbf{Model} & \textbf{Hidden size} \\ \hline
\texttt{DeVries-SL}     & $512$                  \\ \hline
\texttt{DeVries-RL}     & $512$                  \\ \hline
\texttt{GDSE-SL}        & $512$                  \\ \hline
\texttt{GDSE-CL}        & $512$                  \\ \hline
\texttt{GDSE-SL-text}        & $1024$                  \\ \hline
\texttt{GDSE-CL-text}        & $1024$                  \\ \hline
\texttt{GloVe}          & $300$                  \\ \hline
\texttt{ResNet}         & $2048$                 \\ \hline
\end{tabular}

\caption{Summary of hidden state sizes for all the models considered in the attribute prediction evaluation.}
\label{tab:hidden_states}
\end{table}

\begin{table*}[ht!]
\centering
\resizebox{\textwidth}{!}{%
\begin{tabular}{l|c|c|c|c|c|c|c|c|c|c|c|c|}
\cline{2-13}
                                      & \multicolumn{3}{c|}{\textit{\textbf{Abstract}}}                                              & \multicolumn{3}{c|}{\textit{\textbf{Situated-only}}}                                         & \multicolumn{3}{c|}{\textit{\textbf{Abstract+situated}}}                                     & \multicolumn{3}{c|}{\textit{\textbf{Location}}}                                              \\ \hline
\multicolumn{1}{|l|}{\textbf{Models}} & \textbf{F1} & \multicolumn{1}{l|}{\textbf{Precision}} & \multicolumn{1}{l|}{\textbf{Recall}} & \textbf{F1} & \multicolumn{1}{l|}{\textbf{Precision}} & \multicolumn{1}{l|}{\textbf{Recall}} & \textbf{F1} & \multicolumn{1}{l|}{\textbf{Precision}} & \multicolumn{1}{l|}{\textbf{Recall}} & \textbf{F1} & \multicolumn{1}{l|}{\textbf{Precision}} & \multicolumn{1}{l|}{\textbf{Recall}} \\ \hline
\multicolumn{1}{|l|}{DeVries-SL}      & 46.8      & 46.2                                  & 53.4                               & 39.1      & 34.8                                  & 51.2                               & 48.5      & 50.8                                  & 57.8                               & 42.7      & 42.8                                  & 42.9                               \\ \hline
\multicolumn{1}{|l|}{DeVries-RL}      & 45.2      & 44.4                                  & 52.5                               & 38.9      & 34.4                                  & 51                                 & 47.2      & 49.4                                  & 57.4                               & 43.5      & 43.6                                  & 43.6                               \\ \hline
\multicolumn{1}{|l|}{GDSE-SL}         & 59.9      & 59.8                                  & 64.1                               & 47.6      & 44                                    & 58.3                               & 60.1      & 63.8                                  & 65.9                               & 48.3      & 48.6                                  & 48.6                               \\ \hline
\multicolumn{1}{|l|}{GDSE-CL}         & 59.5      & 59.3                                  & 63.6                               & 47.6      & 43.8                                  & 58.6                               & 59.8      & 63.3                                  & 65.6                               & 48.1      & 48.1                                  & 48.6                               \\ \hline
\multicolumn{1}{|l|}{GDSE-SL-text}    & 57        & 56.7                                  & 61.5                               & 45.3      & 41.3                                  & 56.5                               & 57.5      & 60.6                                  & 60.6                               & 46        & 46.1                                  & 46.4                               \\ \hline
\multicolumn{1}{|l|}{GDSE-CL-text}    & 56.9      & 56.9                                  & 61.4                               & 45        & 40.9                                  & 56.4                               & 57.3      & 60.5                                  & 60.5                               & 45        & 45                                    & 45.4                               \\ \hline
\multicolumn{1}{|l|}{GloVe}           & 34.6      & 33.6                                  & 45.9                               & 29.7      & 25.1                                  & 42.1                               & 36.4      & 37.4                                  & 52.9                               & 33.6      & 33.6                                  & 33.7                               \\ \hline
\multicolumn{1}{|l|}{ResNet}          & 24.5      & 24.3                                  & 37.9                               & 31.7      & 27.5                                  & 43.8                               & 27.9      & 30.3                                  & 47.1                               & 43.4      & 43.5                                  & 43.6                               \\ \hline
\multicolumn{1}{|l|}{Random}          & 15.1      & 40.8                                  & 16.3                               & 0.1       & 50.6                                  & 0.1                                & 7.8       & 50.3                                  & 5.4                                & 27.5      & 49.7                                  & 20.3                               \\ \hline
\end{tabular}%
}
\caption{Full set of attribute prediction task metrics. We evaluate F1, Precision and Recall for all the tasks. All the metrics are computed as macro-average.}
\label{tab:full_attribute_prediction}
\end{table*}

We consider the CompGuessWhat?! splits as reference for our experimental evaluation. We generate an hidden state for every successful dialogue and we use the classifier $\Phi$ to predict the target object attributes. We train the classifier $\Phi$ by minimising the binary cross-entropy loss computed between the model prediction and the reference set of attributes. We use ADAM~\cite{kingma2014adam} as optimiser for our training procedure. To prevent overfitting, we perform early stopping on the validation set using the multi-label F1-measure (with threshold $0.75$) as reference metric and we apply a learning rate scheduler to gradually reduce the learning rate. The model training has been implemented using \textit{AllenNLP}~\cite{gardner2018allennlp}. We report the full set of metrics evaluated for this task in Table \ref{tab:full_attribute_prediction}.

\begin{table}[ht!]
\small
\centering
\begin{tabular}{l|l|l|}
\cline{2-3}
& \textbf{\# images} & \textbf{\# games} \\ \hline
\multicolumn{1}{|l|}{\begin{tabular}[c]{@{}l@{}}Near-domain \\ validation\end{tabular}}   & 1208                & 5343              \\ \hline
\multicolumn{1}{|l|}{\begin{tabular}[c]{@{}l@{}}Out-of-domain \\ validation\end{tabular}} & 1306               & 5372              \\ \hline
\multicolumn{1}{|l|}{\begin{tabular}[c]{@{}l@{}}Near-domain\\ test\end{tabular}}          & 3097               & 13836             \\ \hline
\multicolumn{1}{|l|}{\begin{tabular}[c]{@{}l@{}}Out-of-domain\\ test\end{tabular}}        & 3212               & 13300             \\ \hline
\end{tabular}
\caption{Statistics for the \compguesswhat~zero-shot scenario. We provide near-domain and out-of-domain splits using specific \textit{nocaps} images as reference scenes.}
\label{tab:zs_dataset}
\end{table}

For the \texttt{GDSE} models we used a modified version of the code provided by the author via personal correspondence. On the other hand, for the \texttt{DeVries} model, we use the pretrained models and code that is available on the official webpage~\footnote{\url{https://github.com/GuessWhatGame/}}. The \texttt{GloVe} representations have been generated considering the dialogue as a long sequence of tokens and averaging the corresponding word embeddings. We used SpaCy\footnote{\url{https://spacy.io/}} to obtain the representation of the entire dialogue. For the \texttt{ResNet} features we used the ones used by \cite{shekhar2019beyond} based on a pretrained ResNet-152 model~\footnote{\url{https://pytorch.org/docs/master/torchvision/models.html}}.

Models such as GDSE adopt during training a specific constraint on the dialogue length. Particularly, they ignore dialogues having dialogue length greater than $10$. This means that the model is never exposed to dialogues whose length is greater than $10$. So for this family of models, for all those reference dialogues in GuessWhat?! having more than $10$ turns, we consider only the last $10$ turns and we generate the hidden state for the last turn. In general, we also assume that, whenever a model is not able to generate an hidden state representation for a given dialogue, we generate a zero vector. We did not change the behaviour in any way to avoid possible conflicts with the pretrained model. In addition, in this way a model that is not able to generate a representation for the dialogue would be penalised in the evaluation phase.

\subsection{Zero-shot Evaluation Reference Games Generation}\label{appendix:zero_shot_eval}

We define an automatic procedure to generate the set of reference games for the zero-shot learning setup. Specifically, for all the images in \textit{nocaps} validation and test sets we first extract all the bounding boxes that satisfy the following conditions: 1) bounding box area should be greater than 500 pixels; 2) bounding box region should not be \emph{occluded}; 3) bounding box region should not be \emph{truncated}; 4) bounding box should not be associated with human body parts. An additional inclusion condition for the image is that the number of valid bounding boxes should be between $3$ and $20$. This `sanity check' step is inspired by the procedure adopted in the original \guesswhat~dataset \cite{de2017guesswhat} and is used in order to guarantee that the gameplay reference images are not too crowded or composed of really small objects. Finally, we split the valid images in \textit{near-domain} or \textit{out-of-domain}. An image is considered \textit{near-domain} if it contains \emph{at least} an object whose category belongs to MSCOCO; we consider the image \textit{out-of-domain} if it does not contain any MSCOCO category.

\begin{figure*}
    \centering
    \includegraphics[scale=.35,keepaspectratio]{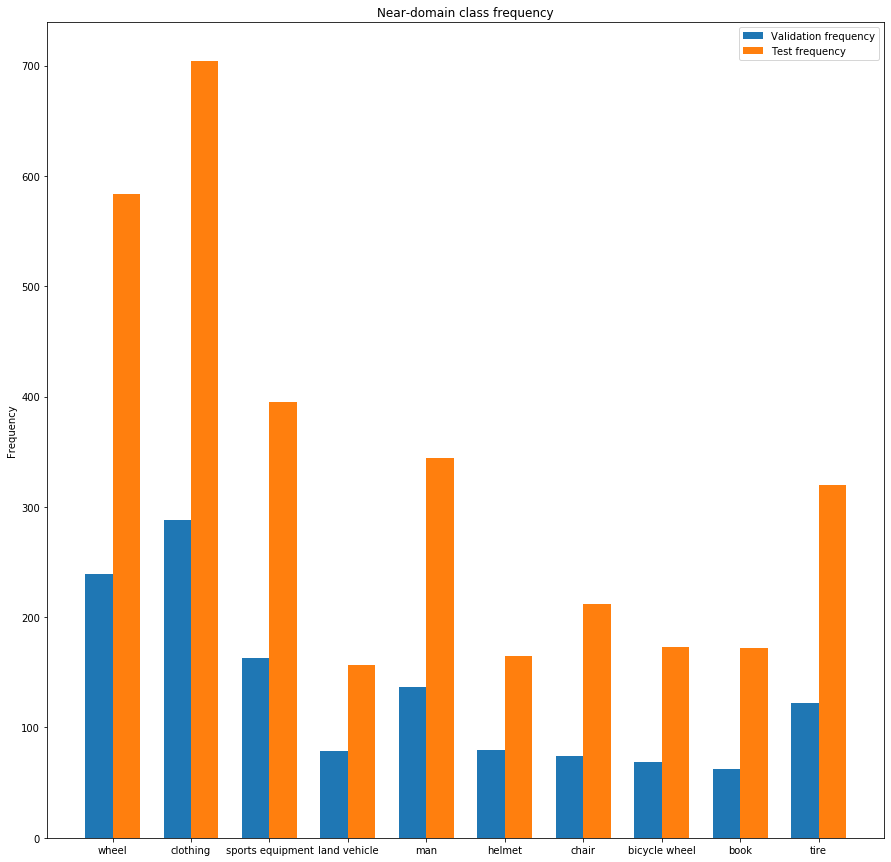}
    \caption{Object category distribution in the near-domain reference set of games.}
    \label{fig:nd_category_distribution}
\end{figure*}

\begin{figure*}
    \centering
    \includegraphics[scale=.35,keepaspectratio]{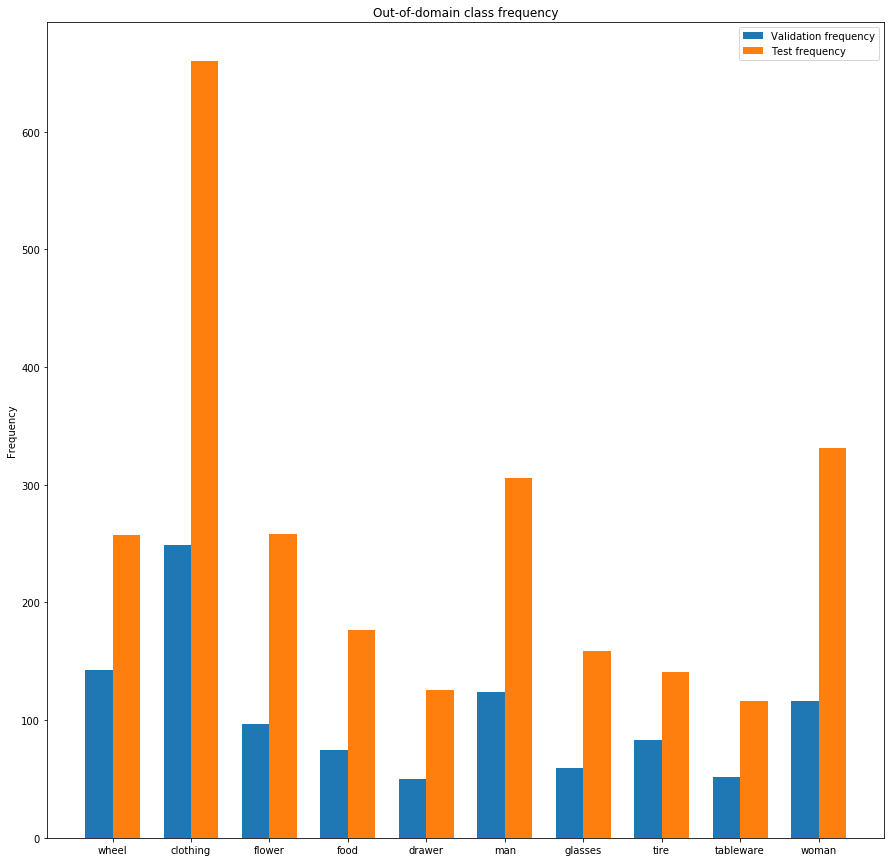}
    \caption{Object category distribution in the out-of-domain reference set of games.}
    \label{fig:od_category_distribution}
\end{figure*}

All the valid images resulting from the sanity check step can be considered as reference scene for the game. In order to define a fair comparison between all the agents, we define a reference set of games by sampling a fixed number of target objects for every image (e.g., $5$ objects). In order to make sure that the sampling procedure is not biased by the frequency of the classes in the dataset, we sample an object according to the inverse of its category frequency in the dataset. 
As a result of this procedure, as shown in Table \ref{tab:zs_dataset}, we generated a dataset of $19,179$ near-domain reference games and $18,672$ out-of-domain reference games. In Figure \ref{fig:nd_category_distribution} and \ref{fig:od_category_distribution} show the object category distribution in the near-domain and out-of-domain reference games, respectively. 

\subsection{Generated Dialogue Evaluation}\label{appendix:dialogue_eval}

In order to evaluate to provide a more fine-grained evaluation of the generated dialogues, we adapt the quality evaluation script presented by \cite{shekhar2019beyond} and extend it with additional metrics. First of all, it relies on a rule-based question classifier that classifies a given question in one of seven classes: 1) super-category (e.g., ``person", ``utensil", etc.), 2) object (e.g., ``car", ``oven", etc.), 3) ``color", 4) ``size", 5) ``texture", 6) ``shape" and ``location". The question classifier is useful to evaluate the dialogue strategy learned by the models. In particular, we look at two types of turn transitions: 1) super-category $\rightarrow$ object/attr, it measures how many times a question with an affirmative answer from the Oracle related to a super-category is followed by either an object or attribute question (where ``attribute" represents the set $\{$color, size, texture, shape and location$\}$; 2) object $\rightarrow$ attr, it measures how many times a question with an affirmative answer from the Oracle related to an object is followed by either an object or attribute question. We compute the \textit{lexical diversity} as the type/token ratio among all games, \textit{question diversity} and the percentage of games with repeated questions. We also evaluate the percentage of dialogue turns involving location questions. Table \ref{tab:nd_zs_quality} and \ref{tab:od_zs_quality} show the results of these analysis for the models \texttt{DeVries} and \texttt{GDSE} analysed in this paper.

\begin{table*}[ht!]
\small
\centering
\begin{tabular}{|l|l|l|l|l|l|l|l|l|}
\hline
\textbf{Model} & \textbf{\begin{tabular}[c]{@{}l@{}}Lexical\\ diversity\end{tabular}} & \textbf{\begin{tabular}[c]{@{}l@{}}Question\\ diversity\end{tabular}} & \textbf{\begin{tabular}[c]{@{}l@{}}\% games \\ repeated \\ questions\end{tabular}} & \textbf{\begin{tabular}[c]{@{}l@{}}Super-cat -\textgreater \\ obj/attr\end{tabular}} & \textbf{\begin{tabular}[c]{@{}l@{}}Object -\textgreater \\ attribute\end{tabular}} & \textbf{\begin{tabular}[c]{@{}l@{}}\% turns\\ location \\ questions\end{tabular}} & \textbf{Vocab. size} & \textbf{Accuracy} \\ \hline
DeVries-SL     & 0.76                       & 44.64                       & 12.54\%                              & 97.33\%                                   & 73\%                                 & 29.34\%                              & 2668                 & 31.33\%           \\ \hline
DeVries-RL     & 0.13                      & 1.77                        & 99.48\%                              & 96.43\%                                   & 98.63\%                                 & 78.07\%                              & 702                  & 43.92\%           \\ \hline
GDSE-SL        & 0.13                      & 6.10                        & 92.38\%                              & 95.60\%                                   & 52.35\%                                 & 15.74\%                              & 862                  & 29.78\%           \\ \hline
GDSE-CL        & 0.17                      & 13.74                       & 66.76\%                              & 99.48\%                                   & 67.25\%                                 & 31.23\%                              & 1260                 & 43.42\%           \\ \hline
\end{tabular}
\caption{Gameplay quality analysis on Near-domain zero-shot reference games.}
\label{tab:nd_zs_quality}
\end{table*}

\begin{table*}[ht!]
\small
\centering
\begin{tabular}{|l|l|l|l|l|l|l|l|l|}
\hline
\textbf{Model} & \textbf{\begin{tabular}[c]{@{}l@{}}Lexical\\ diversity\end{tabular}} & \textbf{\begin{tabular}[c]{@{}l@{}}Question\\ diversity\end{tabular}} & \textbf{\begin{tabular}[c]{@{}l@{}}\% games \\ repeated \\ questions\end{tabular}} & \textbf{\begin{tabular}[c]{@{}l@{}}Super-cat -\textgreater \\ obj/attr\end{tabular}} & \textbf{\begin{tabular}[c]{@{}l@{}}Object -\textgreater \\ attribute\end{tabular}} & \textbf{\begin{tabular}[c]{@{}l@{}}\% turns\\ location \\ questions\end{tabular}} & \textbf{Vocab. size} & \textbf{Accuracy} \\ \hline
DeVries-SL     & 0.83                                                               & 45.86                                                                 & 11.58                                                                              & 97.87\%                                                                              & 76.50\%                                                                          & 29.64\%                                                                             & 2604              & 28.37\%            \\ \hline

DeVries-RL     & 0.24                                                               & 2.96                                                                & 98.49\%                                                                             & 91.83\%                                                                               & 98.58\%                                                                            & 75.84\%                                                                              & 1275             & 38.73\%            \\ \hline

GDSE-SL        & 0.09                                                                & 1.31                                                                & 97.19\%                                                                             & 100\%                                                                             & 67.45\%                                                                             & 7.90\%                                                                             & 519              & 22.32\%            \\ \hline
GDSE-CL        & 0.14                                                              & 7.86                                                                 & 66.32\%                                                                              & 100\%                                                                              & 71.14\%                                                                             & 26.03\%                                                                          & 1002            & 29.83\%            \\ \hline
\end{tabular}
\caption{Gameplay quality analysis on Out-of-domain zero-shot reference games.}
\label{tab:od_zs_quality}
\end{table*}

\end{document}